\documentclass[lettersize,journal]{IEEEtran}
\usepackage{amsmath,amsfonts}
\usepackage{algorithmic}
\usepackage{array}
\usepackage{textcomp}
\usepackage{stfloats}
\usepackage{url}
\usepackage{verbatim}
\usepackage{graphicx}
\usepackage{booktabs}
\usepackage{algorithm}
\usepackage{pifont}
\usepackage{newfloat}
\usepackage[accsupp]{axessibility}  

\usepackage[pagebackref,breaklinks,colorlinks]{hyperref}
\usepackage{orcidlink}
\usepackage{cleveref}
\usepackage{listings}
\usepackage{subcaption}

\newcommand{\etal}{\textit{et al.}}
\crefname{figure}{Fig.}{Figs.}
\crefname{section}{Sec.}{Secs.}
\crefname{table}{Tab.}{Tabs.}
\def\BibTeX{{\rm B\kern-.05em{\sc i\kern-.025em b}\kern-.08em
    T\kern-.1667em\lower.7ex\hbox{E}\kern-.125emX}}
\usepackage{balance}
\begin{document}
\title{IDRetracor: Towards Visual Forensics Against Malicious Face Swapping }
\author{
	Jikang Cheng, Jiaxin Ai, Zhen Han, Chao Liang, Qin Zou, \textit{Senior Member, IEEE},
	Zhongyuan Wang, \textit{Member, IEEE},  and Qian Wang, \textit{Fellow, IEEE} 
 \thanks{Jikang Cheng, Jaxin Ai, Zhen Han, Chao Liang, Qin Zou, and Zhongyuan Wang are with the school of computer science, Wuhan University, Wuhan 430000, China. (e-mail: \{chengjikang, julyai, cliang, qzou\}@whu.edu.cn, hanzhen\_1980@hotmail.com, wzy\_hope@163.com). 
 
 Qian Wang is with the school of cyber science and engineering, Wuhan University, Wuhan 430000, China. (e-mail: qianwang@whu.edu.cn). 

 Corresponding Author: Zhongyuan Wang.
 
 }
}


\maketitle

\begin{abstract}
	The face swapping technique based on deepfake methods poses significant social risks to personal identity security. While numerous deepfake detection methods have been proposed as countermeasures against malicious face swapping, they can only output binary labels (Fake/Real) for distinguishing fake content without reliable and traceable evidence.
	To achieve visual forensics and target face attribution, we propose a novel task named face retracing, which considers retracing the original target face from the given fake one via inverse mapping.
	Toward this goal, we propose an IDRetracor that can retrace arbitrary original target identities from fake faces generated by multiple face swapping methods. Specifically, we first adopt a mapping resolver to perceive the possible solution space of the original target face for the inverse mappings. Then, we propose mapping-aware convolutions to retrace the original target face from the fake one. Such convolutions contain multiple kernels that can be combined under the control of the mapping resolver to tackle different face swapping mappings dynamically. Extensive experiments demonstrate that the IDRetracor exhibits promising retracing performance from both quantitative and qualitative perspectives.
\end{abstract}



\section{Introduction}
Face swapping by deepfake methods \cite{deepfakes, simswap,infoswap,hififace,megaFS,e4s,Styleswap} has achieved remarkable advancements in recent years. Face swapping refers to generating a fake face by replacing the identity (ID) of the target face with the source face ID while preserving the target face attributes (\textit{e.g.}, pose, facial expression, and lighting condition). Such face swapping techniques can pose serious risks to personal identity security. Fake faces that are convincingly realistic can distort public perception and potentially endanger the reputation of the source subjects. Hence, many deepfake detection methods have been proposed to tackle malicious face swapping content. Deepfake detection can output a binary label that indicates whether an image is forged, which can provide judgment results for people to identify deepfake images. However, given that the output of deepfake detection is simply a binary label, there is an inherent lack of traceable and reliable evidence. Meanwhile, identity information and reputation are of extraordinary value to some people (like politicians and celebrities). This situation consequently demonstrates the inadequacy of deepfake detection \cite{seq-deepfake, trace} and thus urges toward visual forensics \cite{ganInv-1} and target face attribution. 

In previous works for visual forensics, \cite{ganInv-1} and the following works \cite{ganInv-2,ganInv-3} introduce the GAN fingerprint to attribute the input forgery content to the source architecture.
Furthermore, Sequential Deepfake Detection (Seq-Deepfake) \cite{seq-deepfake,seq-tifs} is proposed to recover the original face against sequential facial editing. Motivated by these works, we advance beyond mere detection against malicious face swapping content.

\begin{figure}[t]
	\centering
	\includegraphics[width=\linewidth]{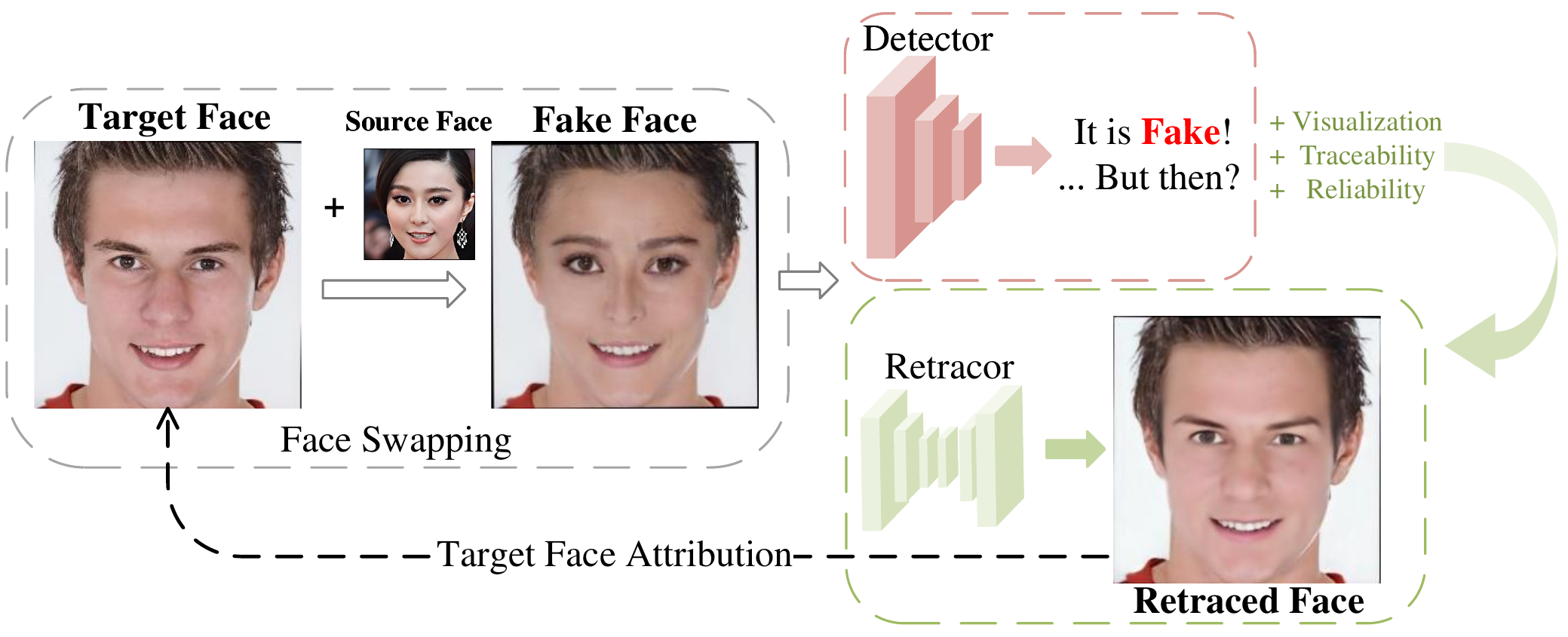}
	\caption{The face retracing task for visual forensics and target face attribution. The deepfake detector can only provide a straightforward Fake/Real label without further information for subsequent actions against the perpetrators. In contrast, the retracor can visualize a retraced face with more traceability and reliability for forensics.  }
    \label{fig:instr1}
\end{figure}
Supposing the forged fake face could be retraced and reconstructed back to its original target face, it would yield more reliable evidence to assess the authenticity of the fake faces for the source subjects who suffered from malicious face swapping. Such a process not only allows for the \textit{identification of the original target face} based on the retraced one, confirming the deception of the fake face image with reliability, but it can also \textit{facilitate tracing the potential perpetrator}, aiding in accountability and rights enforcement. We illustrate the face retracing task in Fig.~\ref{fig:instr1}.


There has been an initial attempt at inversing the original target face from the fake one via DRNet \cite{trace}. However, DRNet requires that the target ID has been seen during training, thereby limiting their ability to handle arbitrary source IDs. Whereas in real-world applications, it is uncommon for the target ID to be included in the training data. Moreover, prior methods are designed to retrace a specific face swapping method, yet the face swapping method used by the malicious perpetrators is usually unknown, which further limits the applicability of DRNet. Consequently, proposing a retracing method with feasibility to resist malicious face swapping remains a considerable challenge.

The feasibility of the retracing task is based on two premises: 1) The fake faces generated by existing deepfake methods inevitably retain implicit target ID information \cite{dect-huang,leakage}. 2) The input fake faces share high ID similarity with the source faces \cite{simswap,infoswap,hififace,e4s}. Then, we posit that it is feasible to solve the original target face based on the given fake face via inverse mapping.
Firstly, we craft a source-fake dataset that contains a large number of various original target IDs and multiple face swapping methods for a designated source subject. This dataset includes abundant information for the trained network to retrace arbitrary target IDs and multiple face swapping methods. However, learning from such a complex dataset poses two major issues: 1) The abundance of original target IDs leads to more challenging mapping relations to be learned. 2) Since different face swapping methods can lead to different mapping rules from target to fake, the form of residual target ID artifacts may vary with each distinct face swapping method, thus leading to alterations in the rules of inverse mapping when reconstructing the original target face. These issues substantially complicate the process of learning the correct inverse mapping solution space for a retracing network.  

In this paper, we present the \textbf{first} practical retracing framework named IDRetracor for visual forensics and target face attribution against malicious face swapping. Unlike previous watermark-based~\cite{fakewatermark} or trigger-based~\cite{faketagger} methods, IDRetracor is the first method that does not require any pre-modification on the original benign data.
Specifically, we introduce the Mapping-aware Convolution that contains multiple groups of convolution kernels to tackle the common and unique mappings of various target IDs and multiple face swapping methods. We then deploy a Mapping Resolver that can resolve the possible mappings of the input face, thereby guiding the mapping-aware convolutions to recombine their kernels dynamically.
Therefore, the recombined kernel can fit different inverse mappings and output retraced faces within the solution space corresponding to the original target face identity. This framework allows us to retrace multiple face swapping methods and arbitrary target IDs, even if the original target ID is not seen during training. 
The contributions of this paper can be summarized as follows:
\begin{itemize}
	\item	We explore the feasibility of retracing the original target face from the fake one, thus pushing the current countermeasures against malicious face swapping towards visual forensics and target face attribution.
	\item	We designed an IDRetracor based on mapping-aware convolutions for retracing, which is trained on a target-fake dataset with a large number of target IDs. IDRetracor can adjust the solution space of inverse mapping dynamically to ensure the retraced face is within the space of the original target face identity.
	\item	Our experimental results clearly exhibit promising performance under the circumstance of arbitrary target IDs and multiple face swapping methods. Qualitatively, the original identity can be intuitively identified from the retraced faces. Quantitatively, the Arcface similarity to the original target exceeds 0.65.
\end{itemize}

\section{Related Work}
\subsection{Deepfake Detection}
Deepfake detection is one of the most popular countermeasures against malicious forgery content. Previous methods \cite{dect-1,dect-2,dect-3,dect-4,dect-5,ucf, leakage,tifs-backbone,tifs-2} for deepfake detection can distinguish the statistical or physics-based artifacts and then recognize the authenticity of visual media. Huang \etal \cite{dect-huang} propose to leverage the implicit identity in the generated fake faces to detect face forgery.  However, these countermeasures have two major problems:\\
1) \textbf{Lack of reliability}:  Rather than furnishing any concrete evidence of forgery content, they offer only a binary label (Fake/Real), which appears less convincing and reliable. \\
2) \textbf{Lack of traceability}: Individuals may be inclined to pursue accountability and rights enforcement against malicious perpetrators, while no further information for tracing is provided by the results of deepfake detection.

These issues strictly constrain the utility of deepfake detection and demonstrate the necessity for an approach that facilitates visual forensics \cite{ganInv-1} and target face attribution. 
\subsection{Visual Forensics against Image Forgery}
Toward visual forensics, there are many efforts \cite{ganInv-1,ganInv-2,ganInv-3, VF1,VF2} focusing on the fake contents generated by the Generative Adversarial Network (GAN). Specifically, they are designed to extract GAN fingerprints from the forgery images for visual forensics and model attribution. Given that a specific GAN architecture has a unique GAN fingerprint, it can be employed to attribute fake images
to the source architecture. Such previous works on GAN fingerprints have demonstrated the severe inadequacy of numerical detection results for forensics, while also validating the feasibility of learning network mappings and the common feature of residual artifacts on the generated images.

Moreover, Shao \textit{et al.} \cite{seq-deepfake} propose Sequential Deepfake Detection (Seq-Deepfake) to recover the original face against sequential facial editing. Xia \textit{et al.}~\cite{seq-tifs} further propose MMNet to enhance the performance of Seq-Deepfake. \textit{Their task is relatively easier since the modification of facial editing is essentially minor than face swapping}. Nevertheless, Seq-Deepfake also demonstrates the prevalent concern about the inadequacy of deepfake detection (\textit{i.e.}, mere binary outcome) and the strong demand for visual forensics.

Focusing on malicious face swapping, Ai \textit{et al.} \cite{trace} propose the Disentangling Reversing Network (DRNet) for face swapping inversion. However, DRNet is limited to retracing target IDs that have been seen during training, whereas in application scenarios, the reversing network often deals with a large number of unknown target IDs. This issue essentially undermines the capability of DRNet to provide substantial value for practical applications. Therefore, DRNet can only be considered a preliminary exploratory attempt.

\section{Methodology}

\begin{figure}[htbp]
    \centering
        \includegraphics[width=0.48\textwidth]{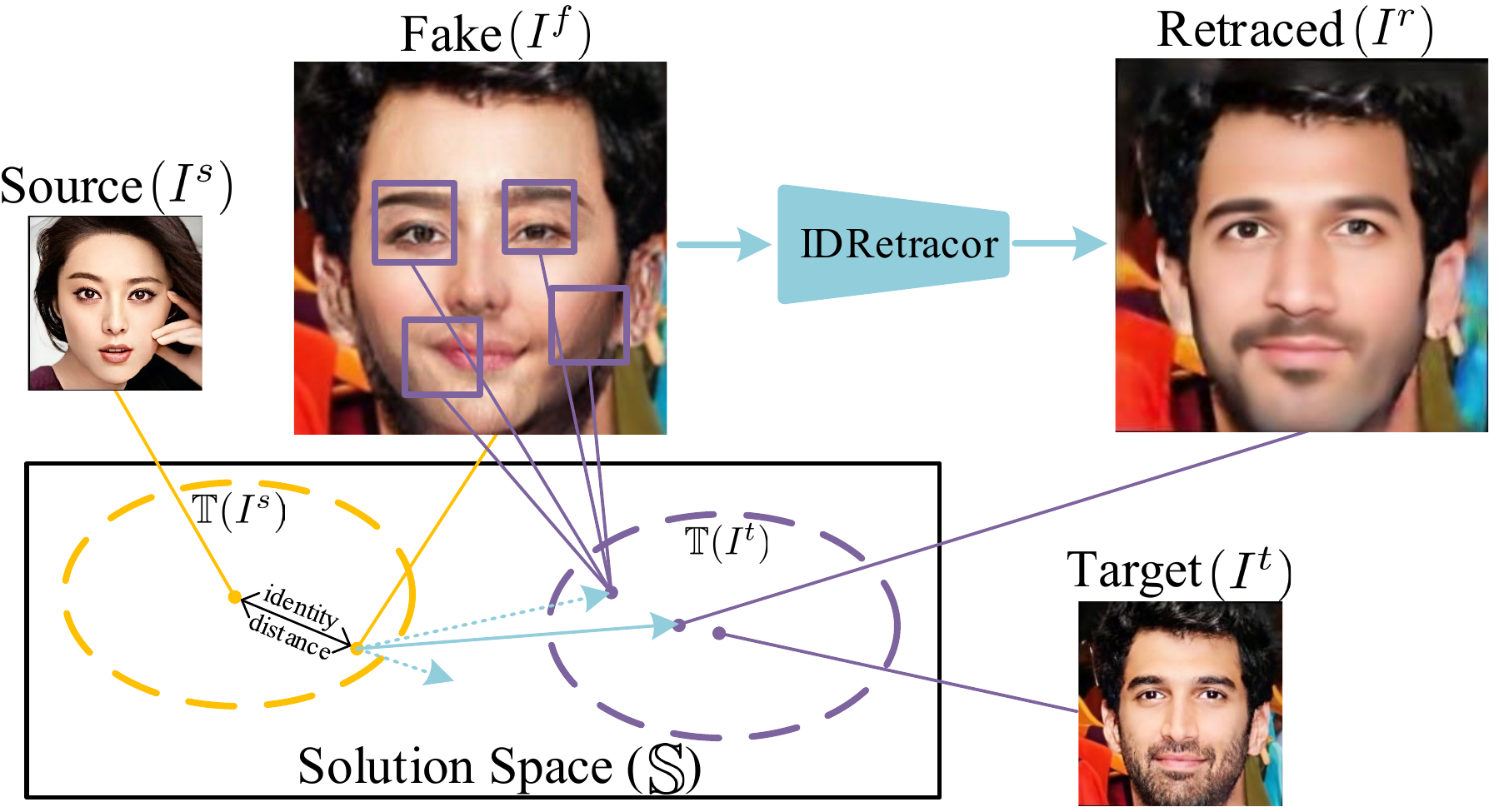}
        \caption{Illustration on the significance of the two premises. The purple squares indicate the residual artifacts in $I^f$. The slimmer blue line denotes the mapping of IDRetracor for solving the original target face. Two thicker blue lines denote the constraint from the two premises for locating the possible solution space. Specifically, the lower line represents convergence towards the direction away from the source ID, while the higher one represents convergence towards the implicit target ID retained in the artifacts.}
        \label{fig:arti}
\end{figure}

\subsection{Premise of Retracing the Target Face}
The process of face swapping can be defined as a mapping:
\begin{equation}
	I^{f}=\mathcal{F}_{i}(I^{s},I^{t}), \label{eq1}
\end{equation}
where $I^{f}$, $I^{s}$,  $I^{t}$ denote the face images of fake, source, target, and $\mathcal{F}_{i}$ denotes a mapping for a specific face swapping method. Our retracing task aims to reconstruct the $I^{t}$ based on the given $I^{f}$, and obtain the final retraced face ($I^{r}$). The complete solution space of $I^r$ can be written as:
\begin{equation}
	\mathbb{S} = \{x | x \in [0, 255]^{c \times h \times w}\},
\end{equation}
where $c$, $h$, and $w$ denote the channel, height, and width of the image, respectively. $\mathbb{T}(I)$ is a subset of $\mathbb{S}$ that includes solutions that share an identical identity with image $I$. We can define $\mathbb{T}(I)$ as:
\begin{equation}
	\mathbb{T}(I) = \{x | Sim(x, I) >\tau\},
\end{equation}
where $Sim(\cdot)$ denotes the face similarity that can be objectively calculated via pre-trained models (\textit{e.g.}, Arcface \cite{arcface} and Cosface \cite{cosface}), and $\tau$ denotes the threshold of similarity score.

The integrated task of face swapping and retracing fundamentally serves as an $a \rightarrow a$ image translation task, the feasibility of which is substantiated by CycleGAN \cite{cyclegan}. Concurrently, the feasibility of the retracing task can be justified from a more intuitive perspective. Suppose that we are managing to retrace the current fake face $I^f_c$, if we traverse through all possible $I^t$ in set $\mathbb{S}$ and generate fake faces $I^f$ using every $I^t$ with the source face, we can establish an assert that it is guaranteed to find a specific $I^t_i$ for which the generated fake face $I^f_i$ is pixel-wise identical to $I^f_c$, and thus determines that $I^t_i$ is the retraced face of $I^f_c$.

However, there is one exceptional \textit{many-to-one} mapping condition that undermines the feasibility: a series of target faces $\{I^{t}_{i},\ldots, I^{t}_{j}\}$ that have completely distinct identities are mapping to a \textit{pixel-wise identical} fake face. Formally, there exists $i,j$ such that $\|\mathcal{F}(I^{t}_i),\mathcal{F}(I^{t}_j)\|_1=0$ while $i\neq j$ and $Sim(I^{t}_i,I^{t}_j)<\tau$. 
In this condition, it becomes inherently infeasible to locate the specific $\mathbb{T}(I^t)$ since its inverse mapping becomes a \textit{one-to-many} mapping that the corresponding original $I^t$ cannot be specified from $\{I^{t}_{i},\ldots,I^{t}_{j}\}$. Consequently,  given that the identities within $\{I^{t}_{i},\ldots,I^{t}_{j}\}$ are distinct, a misguided choice in inverse mapping will culminate in an incorrect identity being attributed to the retraced face. Such a condition could arise in the scenario of the face swapping method being flawless and leaving no residual ID information in the fake face. Alternatively, it could also happen in the scenario in which the swapping method is too poor that the generated fake face shares little identity similarity with the source face. Therefore, the feasibility of the retracing task cannot be fundamentally established without excluding the two mentioned scenarios that lead to the many-to-one mappings with distinct target identities.


Fortunately, a flawless face swapping method that perfectly swaps the identity and leaves no implicit identity information may not exist currently. Specifically, the definition of target and source identity for the existing face swapping methods appears vague. That is, the existing face swapping methods deploy distinct deep learning networks to extract the identity information for training, which is a black-box extracting process and cannot strictly and consistently represent the real abstracted identity. Consequently, as demonstrated by Huang \etal \cite{dect-huang} and \cite{leakage}, the implicit identity information related to the target face is inevitably retained in the fake one. For the second scenario, a poorly forged fake face cannot pose a threat to the source subject, hence there is no necessity for retracing the poorly forged faces. Therefore, we assume two premises for the retracing task: 
\begin{description} 
	\item[\textbf{Premise 1.}] \hspace{1cm} $I^f$ retains implicit identity related to $I^t$. 
	\item[\textbf{Premise 2.}] \hspace{1cm} $I^f \in \mathbb{T}(I^s)$.
\end{description}

As shown in Fig.~\ref{fig:arti}, these premises contribute to further constraining the solution space for the reconstructed $I^{r}$, making it possible to fall within $\mathbb{T}(I^t)$. In summary, within the scope of this paper, we study the face swapping methods that \textit{empirically obey these two premises}, which include all existing face swapping methods to the best of our knowledge. While under conditions where these premises remain intact, the retracing task for visual forensics is theoretically feasible.

\begin{figure*}[t]
	\centering
	\includegraphics[width=0.85\linewidth]{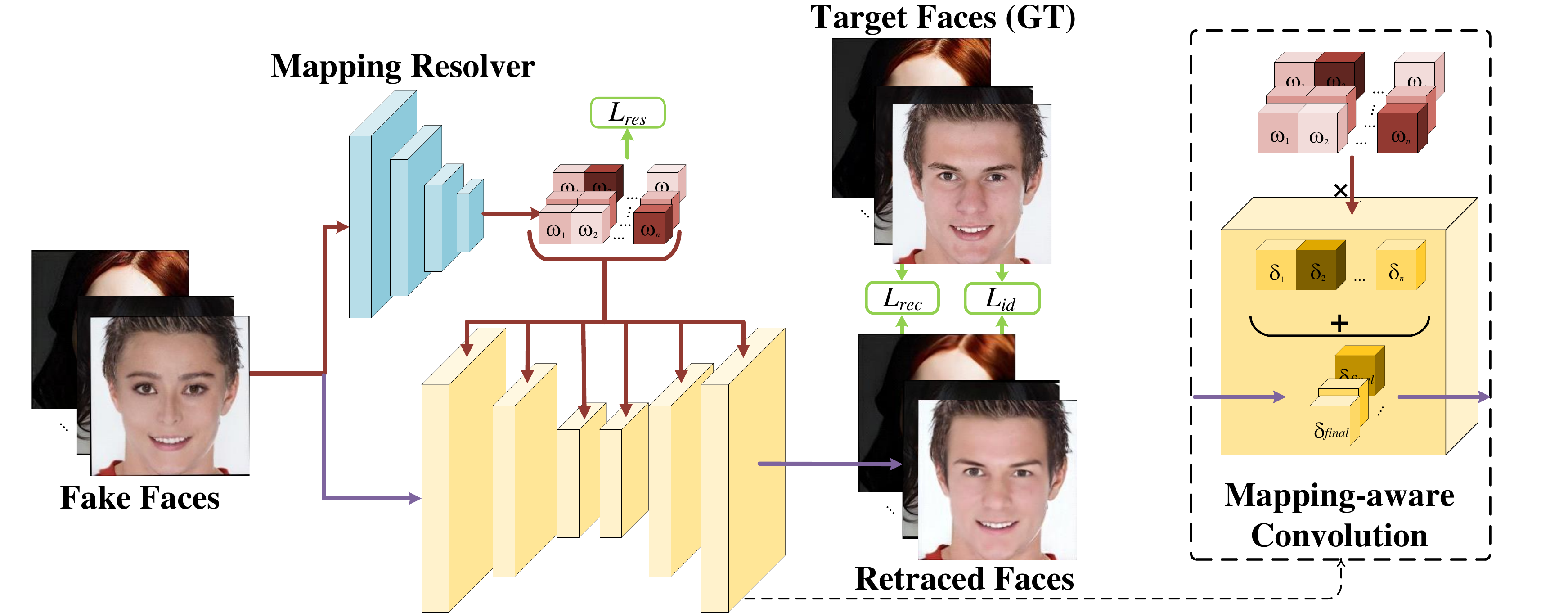}
	\caption{The architecture of the proposed IDRetracor for retracing the original target faces. The mapping resolver can generate various weight vectors according to the input fake faces. These vectors then control the mapping-aware convolutions to recombine a kernel specifically for the input fake faces. Finally, the different recombined kernels process their corresponding fake faces and produce the final retraced faces.  }
 \vspace{-0.3cm}
	\label{fig:mainarch}
\end{figure*}

\begin{figure}[htbp]
        \includegraphics[width=0.48\textwidth]{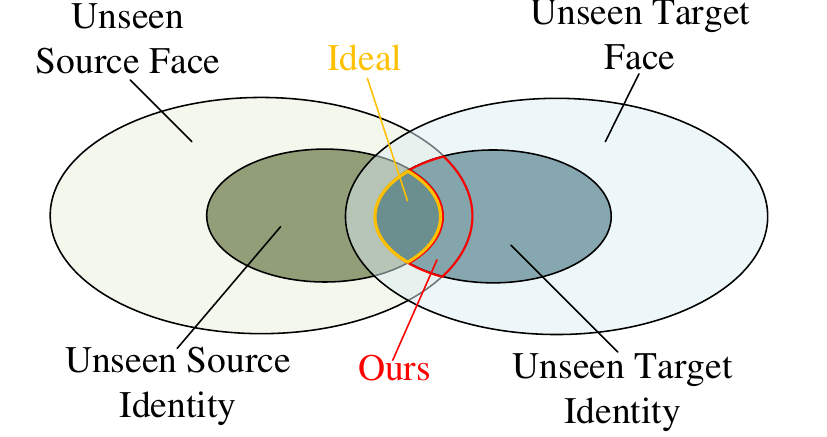}
        \caption{The applicability scope of the retracing task. The region encircled by a yellow line represents the ideal application scope, wherein neither the target nor the source identity is seen during training. The scope of the proposed method is indicated by the red-line region, where the source identity for testing is seen during the training phase, while the specific source face and the target identity are kept entirely unseen.}
        \label{fig:seennunseen}
\end{figure}
\subsection{Constructing an IDRetracor}
Here, we extend the discussion from the feasibility to the practical application, and the applicability scope of the proposed method is illustrated in Fig.~\ref{fig:seennunseen}. Notably, the ideal applicability scope conflicts with Premise 2: Without prior experience of the source identity, it is inherently impossible to verify the authenticity of Premise 2 and thus leads to its ineffectiveness. Hence, the proposed method successfully achieves both feasibility and the broadest possible scope of applicability.

To ensure tackling malicious face swapping in real-world scenarios comprehensively, it is necessary that arbitrary target and multiple face swapping methods can be correctly retraced. We consider utilizing a neural network to solve for $I^t$ in Eq.~\ref{eq1}. It is obvious that the network should learn the mapping of $\mathcal{F}$ and the feature of $I^s$ to successfully reconstruct $I^t$. And under the premise of identity residuals, we may achieve visual forensics and target face attribution.

First, we craft a target-fake dataset for a specific source subject, which incorporates a large number of different target IDs and multiple face swapping methods. The abundant information included in the dataset allows the network to learn different mappings and extend its generalization to arbitrary target faces. 
Notably, we further integrate real source faces as the source-source sample pairs. By training on these sample pairs, the network is capable of producing output that is nearly identical to the input when it is a real one. This implementation allows the network to handle the input images comprehensively, that is, a fake face yields a retraced output, while a real face leads to an output that closely approximates the input.

However, learning on such a complex dataset is a challenging endeavor. Specifically, retracing arbitrary target faces requires learning common features of $I^s$ across a large variety of target-fake image pairs, which intensifies the complexity of network optimization. Meanwhile, learning multiple face swapping methods involves the learning of diverse mappings represented by $\{\mathcal{F}_1,\ldots,\mathcal{F}_n\}$, as opposed to solely focusing on a single $\mathcal{F}_i$.

To address these issues, we propose IDRetracor to retrace the original target faces. The architecture of IDRetracor is shown in Fig.~\ref{fig:mainarch}. We use mapping-aware convolutions to tackle complex mappings and introduce a mapping resolver to dynamically resolve the potential solution space of the current input. The resolved results will guide the mapping-aware convolutions to retrace the solution and ensure $I^r$ falls within $\mathbb{T}(I^t)$. 

\paragraph{Mapping-aware Convolution.}
For $\{\mathcal{F}_1,\ldots,\mathcal{F}_n\}$ and $I^s$, there exist common mappings and features as well as unique ones. Consequently, employing one part of parameters to discern the common mappings and features, while adaptively assigning different parameters for the unique ones, can achieve the effect of fitting the current $\mathcal{F}_i$ through parameter recombination. 

Inspired by \cite{dynamic}, we introduce mapping-aware convolution that contains $n$ groups of convolution kernels, denoted by $C_m=\{\delta_1,\ldots,\delta_n\}$. $C_m$ can learn the common contents and unique ones by different kernels. During the retracing process, the kernels will be adaptively recombined to different final kernels and they will be employed for different input images. The recombination is guided by a weight vector $\Omega=\{\omega_1,\ldots,\omega_n\}$, which is produced by the mapping resolver. Subsequently, the final kernels for retracing can be written as $\delta_{final}=\sum_{i=1}^n\omega_i\delta_i$, 
where $i$ denotes the index of the convolution kernel in our mapping-aware convolution. 
Hence, through recombination, the generated $\delta_{final}$ can tackle both common and unique parts dynamically, thereby fitting to the corresponding $\mathcal{F}_i$ and produce $I^r \in \mathbb{T}(I^t)$.

\paragraph{Mapping Resolver.}
To resolve the potential mappings of the input images, we deploy a mapping resolver, denoted by $\mathcal{R}$. The mapping resolver aims to learn how to resolve different mappings and represent the resolved results through a weight vector $\Omega=\{\omega_1,\ldots,\omega_n\}$, which can be utilized to guide the subsequent mapping-aware convolutions. 

Given that different face swapping methods are considered different mappings, $\mathcal{R}$ could be interpreted as a mapping classifier to some extent. Consequently, the mapping resolver can utilize a standard classifier as its backbone architecture, for which we introduce ResNet18 \cite{resnet}. Then, we employ the Cross-Entropy Loss, a widely adopted constraint in classification tasks, to regulate the resolver. Additionally, inspired by label smoothing \cite{labelsmoothing}, we introduce fusion regularization to promote the generation of non-binary weight, which brings three advantages to the IDRetracor: 1) Prevents the IDRetracor from regressing into an ensemble of different single-mapping models. 2) Encourages processing common content with sharing parameters while processing unique content with distinct parameters. 3) Provides the resolver with adaptability and generalization to some anomalous training image pairs, which inevitably occurs during crafting such a large dataset. Finally, the resolving loss to constrain the resolver can be written as:
\begin{equation}
	L_{res} = -[\theta \log(\omega_{c}) + \frac{1-\theta}{n-1} \sum_{i=1,i\neq c}^{n} \log(\omega_{i})] ,
\end{equation}
where $c$ denotes the index of the current face swapping method and $\theta$ denotes the one-hot value after fusion regularization. Additionally, $\theta$ is set to $\frac{1}{n}$ when the input training sample is the source-source sample pair, which allows each group of convolutions to evenly learn the principles for processing real faces.

\paragraph{Overall Architecture.}
Firstly, given an input fake face $I^f$, the IDRetracor can resolve the mappings of $I^f$ and produce a weight vector via mapping resolver: $\Omega=\mathcal{R}(I^f;\phi_R)$, where $\mathcal{R}$ is parameterized by $\phi_R$. Then, a three-layer UNet \cite{UNet} is employed as our reconstruction network for the original target face retracing. Notably, the UNet we deployed is a modified version, denoted by $\mathcal{N}$, where all vanilla convolutions are replaced by the proposed mapping-aware convolutions. During retracing, $\mathcal{N}$ is guided by $\Omega$ and recombines its kernels dynamically for the final retraced face $I^r=\mathcal{N}(I^f,\Omega;\phi_N)$, where $\mathcal{N}$ is parameterized by $\phi_N$.
To constrain $\mathcal{N}$, we introduce a pixel-wise reconstruction loss $L_{rec}$, defined as follows:
\begin{equation}
	L_{rec} = ||I^r - I^t||_2 ,
	\label{loss1}
\end{equation}
where $||\cdot||_2$ denotes the Euclidean distance. 

Moreover, since we aim to reconstruct the target face that has an identical ID to the original one, we implement $L_{id}$ to constrain the identity information of the retraced face. $L_{id}$ is calculated based on the cosine similarity:
\begin{equation}
	L_{id}= 1- \frac{v^r v^t}{||v^r||_2 ||v^t||_2} ,
\end{equation}
where $v^r$ and $v^t$ denote the feature vectors of $I^r$ and $I^t$ extracted by Arcface recognition model \cite{arcface}. Despite a complete pixel-wise identicalness (\textit{i.e.}, solely optimizing $L_{rec}$ to zero without introducing $L_{id}$) suggesting a complete identity alignment, introducing $L_{id}$ remains significant. This is due to the fact that $L_{rec}$ cannot be always effectively minimized while relaxing the pixel-wise constraint via $L_{id}$ may lead to superior alignment with high identity similarity.

Overall, the total loss function $L_{total}$ that is used to train the IDRetracor can be written as:
\begin{equation}
	L_{total}=L_{rec} + \alpha L_{id} + \gamma L_{res} ,
\end{equation}
where $\alpha$ and $\gamma$ denote the trade-off parameters for $L_{id}$ and $L_{res}$, respectively.

\begin{figure}[t]
    \centering
    \begin{subfigure}[t]{0.15\textwidth}
        \centering
        \includegraphics[width=\textwidth]{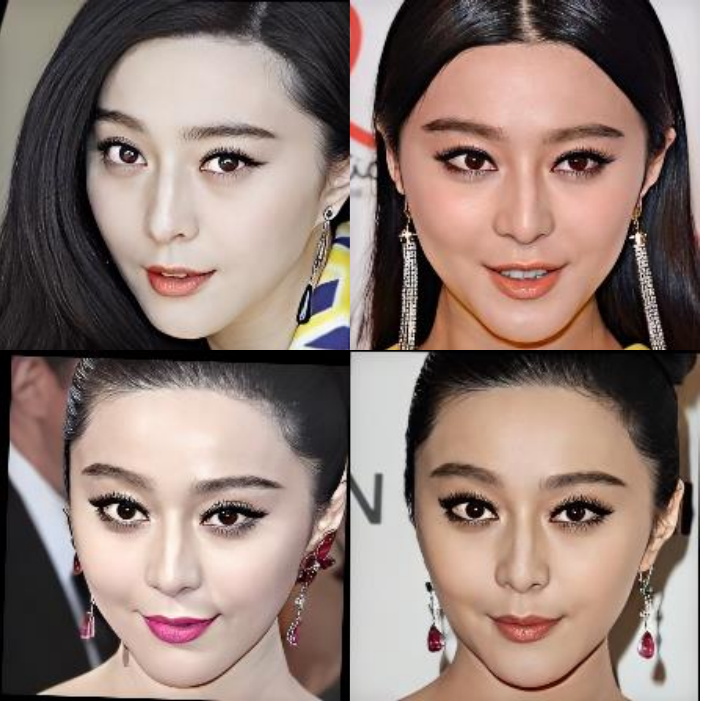}
        \caption{Fan}\label{a}
    \end{subfigure}
    \hfill
    \begin{subfigure}[t]{0.15\textwidth}
        \centering
        \includegraphics[width=\textwidth]{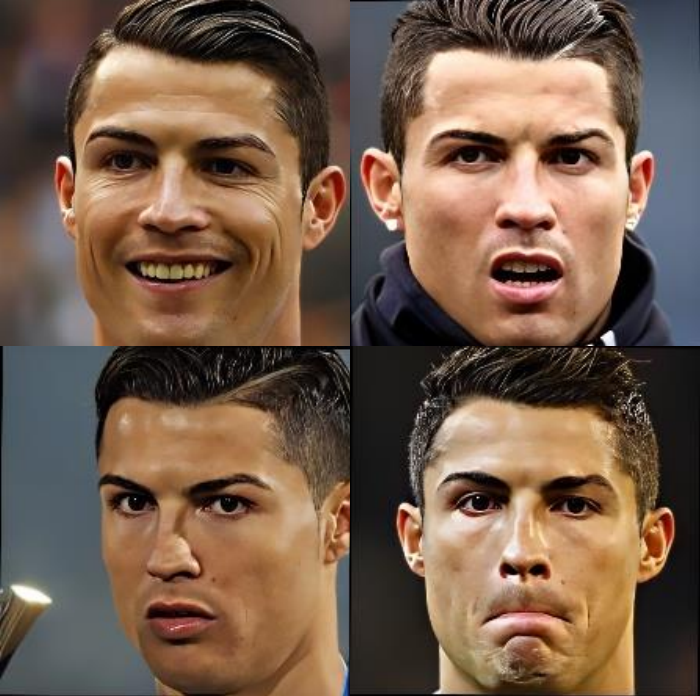}
        \caption{Ronaldo}\label{b}
    \end{subfigure}
    \hfill
    \begin{subfigure}[t]{0.15\textwidth}
        \centering
        \includegraphics[width=\textwidth]{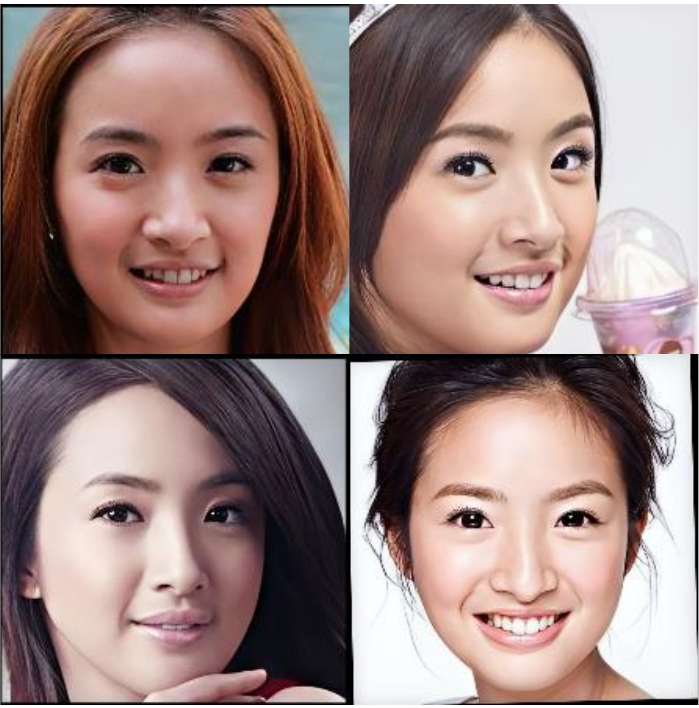}
        \caption{Lin}\label{c}
    \end{subfigure}
    \caption{Examples of source faces from VGGFace2 to craft our target-fake dataset.} \label{fig:source}
\end{figure}
\begin{figure}[t]
	\vspace{0cm}
	\centering
	\includegraphics[width=1\linewidth]{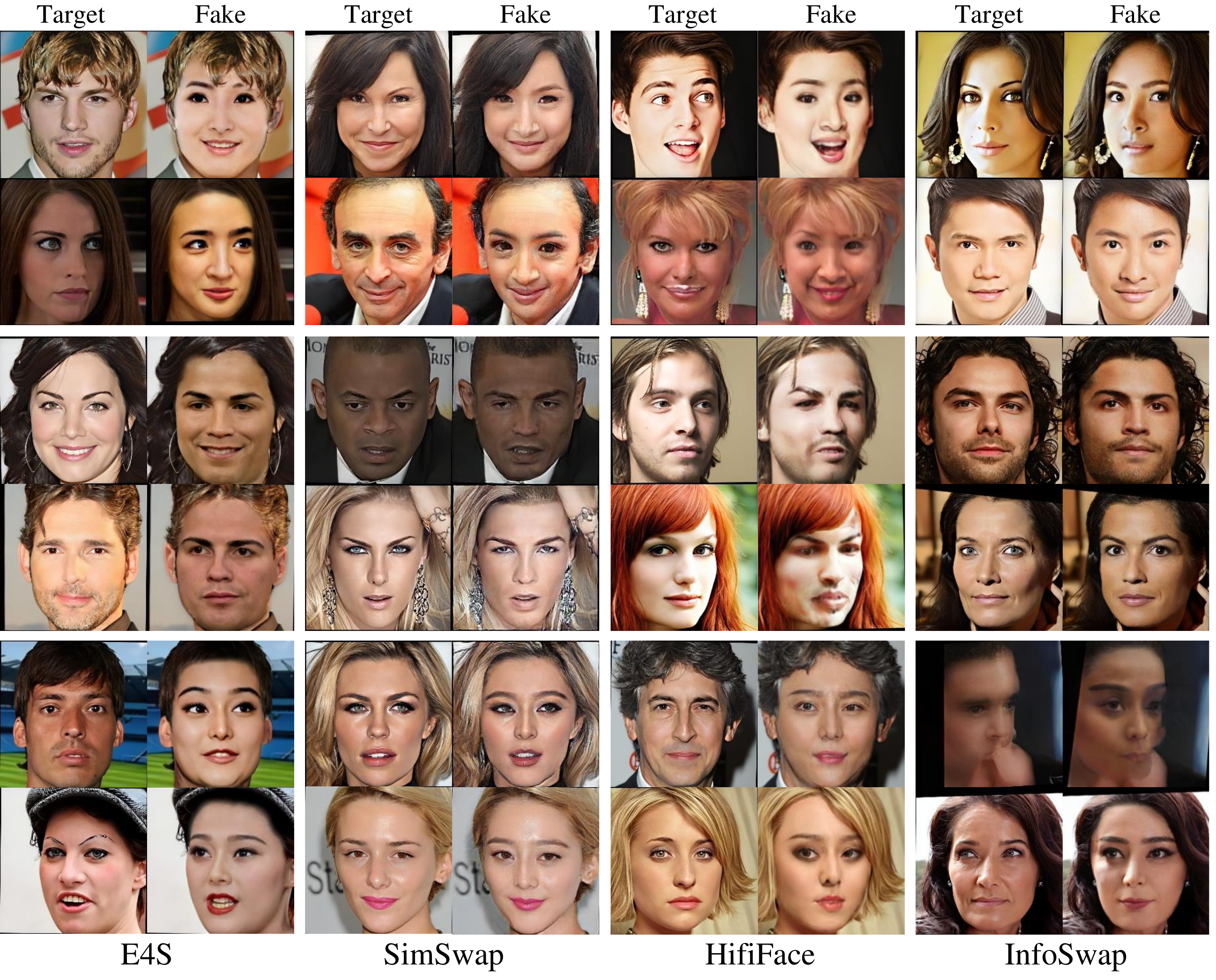}
	\caption{Examples of target-fake sample pairs from the crafted dataset.}
	\label{fig:dataset}
\end{figure}
\section{Experiments}
\subsection{Implementation Details} \label{imp}
\paragraph{Dataset.}
The most widely adopted public datasets for face swapping are Celeb-DF-v2 \cite{Celeb} and FaceForensics++ \cite{Faceforensics}. However, these datasets do not align with the demands of our task due to two primary reasons: 1) They offer an inadequate number of target IDs that pair with one specific source ID. This inadequacy severely limits the generalization of a retracing network to arbitrary target IDs. 2) The face swapping methods employed to generate these datasets remain undisclosed, resulting in the dysfunction of $L_{res}$. 

Consequently, we craft datasets for different source subjects based on VGGFace2 \cite{vggface2}. The dataset includes four face swapping methods reproduced following their official codes, that is, SimSwap \cite{simswap}, InfoSwap \cite{infoswap}, HifiFace \cite{hififace}, and E4S \cite{e4s}. For each face swapping method, we pair the specific source subject with various target IDs from VggFace2. Given that each ID includes multiple faces with distinct face attributes, we pair each different face and finally obtain 150000 training samples and 10000 testing samples for each source subject and face swapping method. The training set for each face swapping method contains 150,000 target-fake sample pairs and 4,636 target IDs, while the test set contains 10,000 target-fake sample pairs and 1,301 target IDs. Following Fig.~\ref{fig:seennunseen}, the target IDs and specific source faces for generating fake faces are completely unrepeated between the testing and training sets. We introduce 180 source-source sample pairs where 140 for training and 40 for testing. All faces are aligned and cropped to 224$\times$224. Examples of the fake faces in the dataset are presented in Fig. \ref{fig:dataset}. Considering the inconsistent performances of every face swapping method, they are not always capable of generating optimal fake faces for training (as shown in row 5, column 4). Such suboptimal data, which can be perceived as noise, may obstruct the learning process of the network. By introducing label smoothing \cite{labelsmoothing} into IDRetracor, we can to some extent mitigate this problem by preventing the model from overfitting to noisy data.

Although this dataset can be easily reproduced following the details we provided, we are still willing to release the dataset to the public subsequent to the publication of this paper. In this paper, the experimental results on the dataset of Bingbing Fan (Fan) will serve as the representative for comparisons and validations (See Fig.~\ref{a} for Fan).

\paragraph{Baselines.}
Since we introduced the \textbf{first} applicable retracing method, there\textit{ exists no previous baseline} for comparison. To further investigate the retracing task, we trained two more models as described below. 1) \textbf{\textit{VU-S}}: Vanilla UNet \cite{UNet} trained with samples generated by a Single face swapping method. In our experiments, we manually deploy the model corresponding to the face swapping method that generates the input fake images. 2) \textbf{\textit{VU-M}}:  Vanilla UNet trained with samples generated by Multiple face swapping methods. UNet backbone is selected to experimentally verify the feasibility of the retracing task. It can be optionally replaced by other image-to-image translation backbones. All methods are trained in an end-to-end manner with 150 epochs and batch size 32. We adopt Adam \cite{adam} optimizer with $\beta_1=0$ and $\beta_2=0.99$ and the learning rate is set to 0.003. The hyper-parameters for IDRetracor are set as $\theta=0.9$, $n=4$, $\alpha=0.01$, and $\gamma=10$. All experiments are implemented in PyTorch on one NVIDIA Tesla A100 GPU. 
\paragraph{Metrics.}
The quantitative evaluations are performed in terms of two metrics: ID similarity and ID retrieval. For ID similarity, we employ two face recognition models (\textit{i.e.}, Arcface \cite{arcface} and Cosface \cite{cosface}) to extract the feature vectors of identity, denoted by \textit{ID Sim. (arc/cos)}. ID retrieval is employed to measure the model capability of retracing the potential perpetrators. To compute ID retrieval, we first randomly select 1000 target faces with different IDs for each face swapping method from the testing set and extract their feature vectors via the Arcface model. Then, we retrace the fake faces that are paired with the selected target faces. Finally, the ID retrieval is measured as the top-1 and top-5 (\textit{ID Ret. (t1/t5)}) matching rates (\%) of the retraced faces and their corresponding target faces.
\begin{figure*}[h]
	\centering
	\includegraphics[width=1\linewidth]{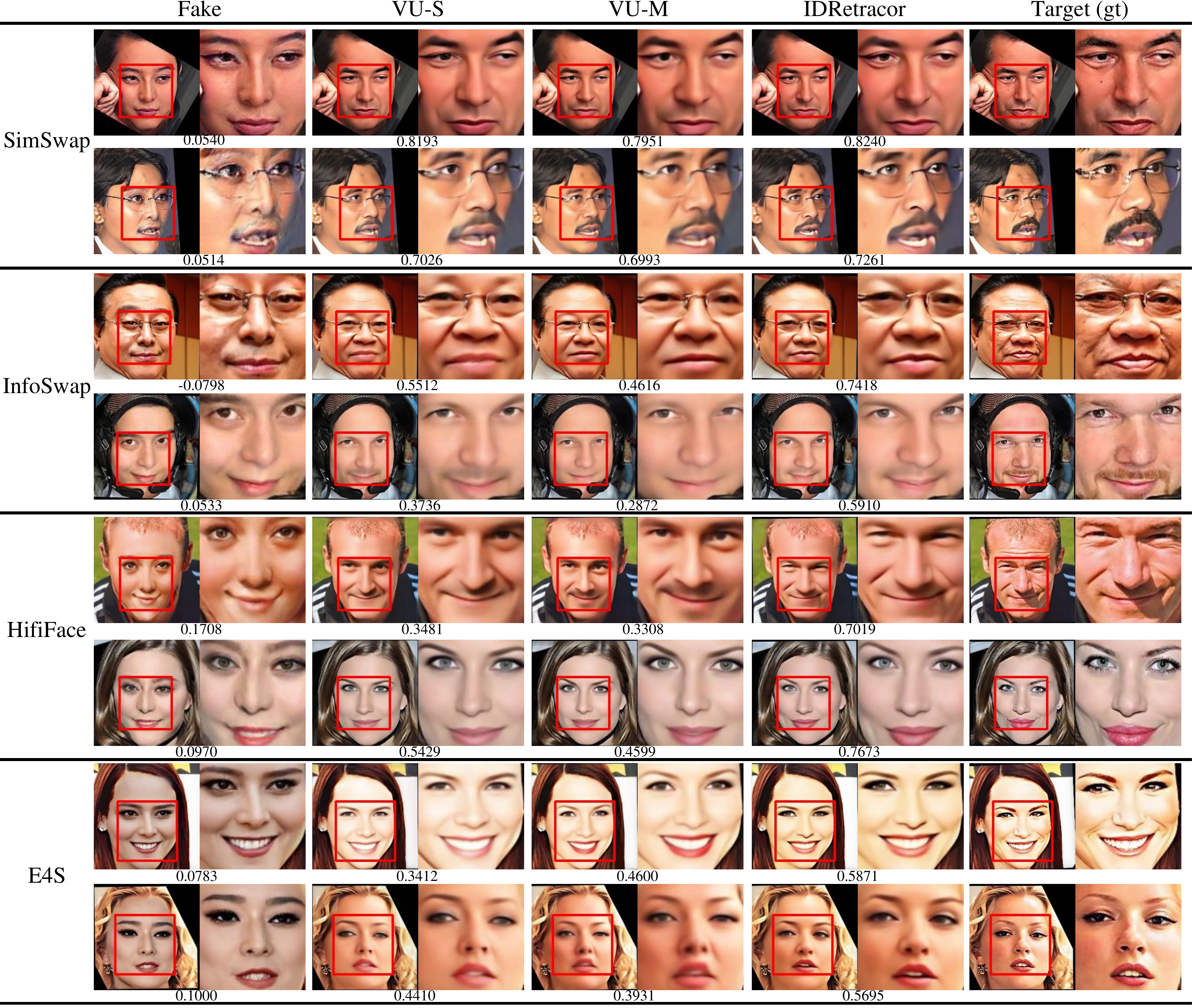}
	\caption{The qualitative results of different retracing models against multiple face swapping methods. The values under each face image are their Arcface Similarity with the target faces. Zoom in for better illustration.}
	\label{fig:maincomparison}
\end{figure*}
\subsection{Retracing Performance}
\begin{figure*}[t]
	\centering
	\includegraphics[width=\linewidth]{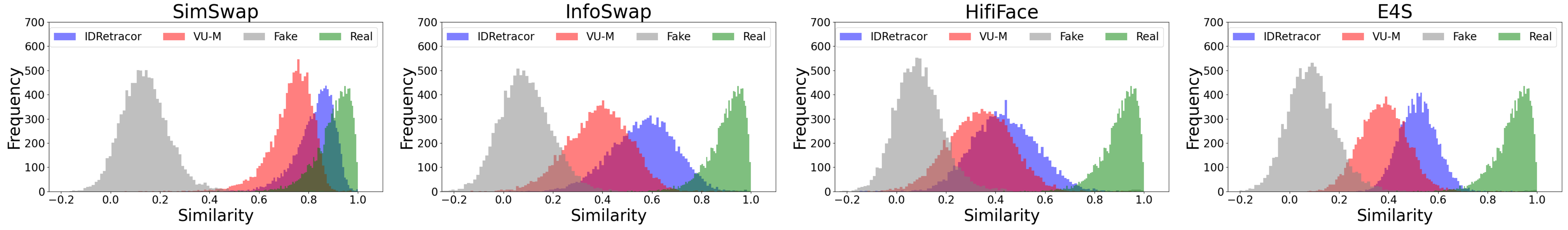}
	\caption{The Arcface Similarity distributions for retracing the fake faces generated by different face swapping methods. We use the fake/retraced faces with their corresponding real faces to calculate their Arcface similarity distribution. We use the real faces that have the same identity to calculate the Real distribution.}
	\label{fig:distribution}
\end{figure*}
\begin{table}[t]
    \small
	\centering
	\captionof{table}{ The quantitative results of retracing models}\label{tab:main}
	\begin{tabular}{lccc}
		\hline
		Method & ID Sim. (arc/cos) & ID Ret. (t1/t5)  & Cross-FS\\  \hline 
		Fake       & 0.1102/0.1397  &   5.53/10.35 &     --   \\  \hline 
		VU-S       & 0.5335/0.6570  &     72.70/85.13  &  \ding{55}  \\ 
		VU-M       & 0.4848/0.6267  &  72.63/83.18  & \ding{51}\\ 
		IDRetracor & \textbf{0.6520/0.7011}  &    \textbf{85.70/96.00} & \ding{51} \\ \hline
	\end{tabular}
\end{table}
\paragraph{Quantitative Results.}
As shown in Tab.~\ref{tab:main}, we present the ID Sim. (arc/cos) and ID Ret. (t1/t5) between the retraced faces and the original target faces. Cross-FS denotes the ability to cross multiple face swapping methods. Notably, the proposed IDRetracor outperforms VU-M by over 20\% and is even higher than VU-S, which is not Cross-FS. Such performances reveal the superiority of the proposed mapping-aware convolutions, that is, they can adapt the mappings of current input dynamically and retrace within the precise solution space of the original target face. 

For a better illustration, we visualize the distribution of Arcface similarity of models that are Cross-FS on the testing set (see Fig.~\ref{fig:distribution}). The distributions demonstrate that the retraced faces can be used as reliable evidence for visual forensics and target face attribution.

\paragraph{Qualitative Results.}
As shown in Fig.~\ref{fig:maincomparison}, we present the retraced faces of VU-S, VU-M, and IDRetracor against four \textit{state-of-the-art} face swapping methods \cite{simswap, infoswap,hififace,e4s}. Considering that SimSwap modifies minor content of the target face during face swapping, the process of retracing becomes considerably easier. Consequently, all three models exhibit strong performance when retracing fake faces produced by SimSwap, while IDRetracor demonstrates a heightened capability of detail restoration. For HifiFace and InfoSwap, which erase more target ID information from the original target faces, IDRetracor can produce retraced faces that are more consistent with the original ID of the target faces. As for the hardest task E4S, the target feature residuals including skin color and facial expression are excessively removed from their original target faces. Despite the immense challenge of fully reconstructing details (\textit{e.g.}, makeup and wrinkles), IDRetracor can still produce retraced faces with the highest identity similarity with the original target faces.

\begin{figure}
    \centering
    \includegraphics[width=0.9\linewidth]{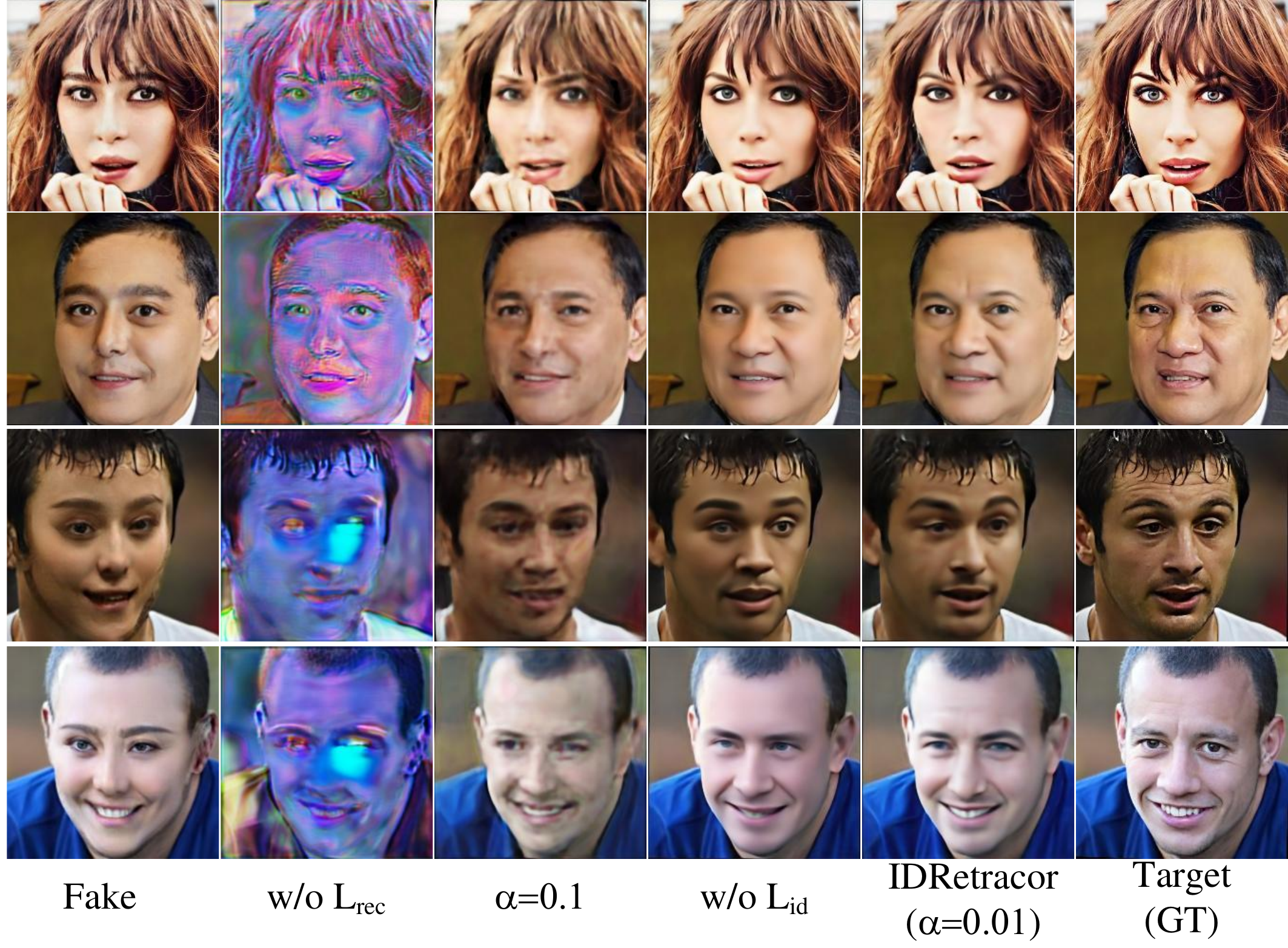}
    \vspace{-0.1cm}
    \caption{The impact of $L_{id}$ weights.} \label{fig:idloss}
\end{figure}


\subsection{Ablation Study}
\paragraph{Effect of Identity Loss.}
To validate the effectiveness of intensity loss ($L_{id}$), we present the retraced results with different $\alpha$ in Fig.~\ref{fig:idloss}. Without $L_{id}$, the retraced results may contain some erroneous eye details (see the third row), while other facial features appear overly smooth (see the first and second rows). This could be due to the challenging conditions in the solution space, where solely constrained by $L_{rec}$ causes the model to lean towards creating a generic face, rather than a result more closely aligned with the original target face ID. In contrast, with the constraint of $L_{id}$, the model tends to supplement details that contribute to enhancing identity similarity. We also conduct additional experiments when $\alpha$ is large (\textit{i.e.}, $\alpha=0.1$) and solely using $L_{id}$.
The results demonstrate that $L_{rec}$ primarily functions to preserve the contour and color of the retraced results, while $L_{id}$ acts as an auxiliary constraint to enhance the details.

\begin{figure}[hb]
\centering
    \includegraphics[width=0.7\linewidth]{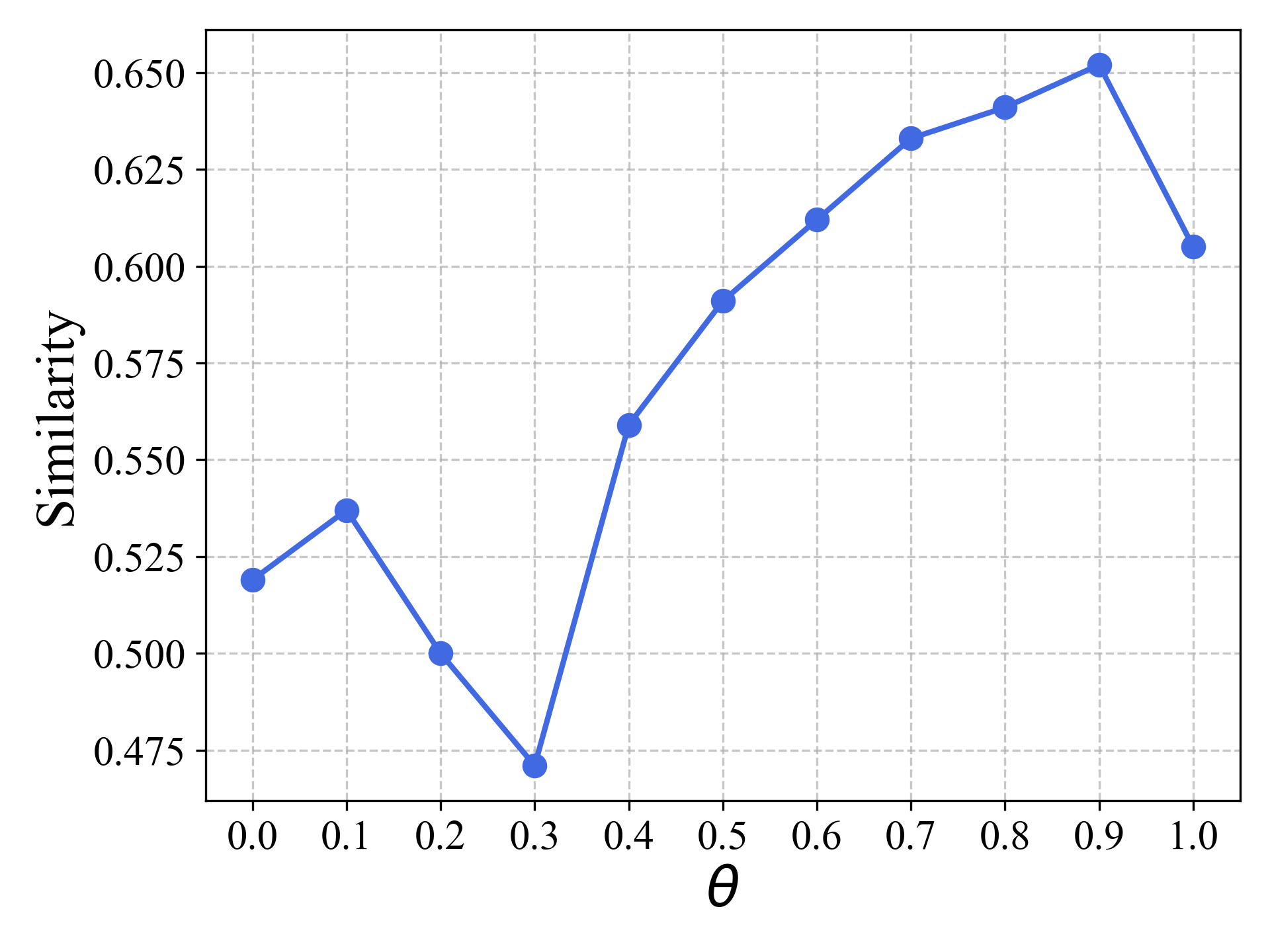}
    \captionof{figure}{The impact of $\theta$ on fusion regularization.}
    \label{fig:sftlabel}
\end{figure}
\paragraph{Effect of Fusion Regularization.}
Here, we analyze and validate the effectiveness of fusion regularization through experiments. As shown in Fig.~\ref{fig:sftlabel}, we measure the average Arcface similarities under different theta values on the testing set. 
When $\theta \in [0.0,0.4]$, the model degrades to a manner of mixed training like VU-M, where the model has a poor resolving ability for shared content and unique content. The worst performance occurs when $\theta=0.3$, as the generated weight vector is almost evenly distributed. When $\theta=1.0$, that is, when fusion regularization is not deployed, the model degrades to a framework that first identifies the face swapping method, and then processes the fake faces with the corresponding VU-S. Therefore, we set $\theta=0.9$ where the IDRetracor exhibits the best performance.

\begin{table}
\centering
\captionof{table}{The performance on the testing set of seen target identities, seen source faces, and unseen data. } \label{tab:inout}
\begin{tabular}{l@{\hspace{0.6cm}}c@{\hspace{0.6cm}}c}
\hline
  & ID Sim. (arc/cos) & ID Ret. (t1/t5)\\  \hline 
Unseen & 0.6520/0.7011  &  85.70/96.00       \\ \hline
Seen-SF & 0.6608/0.7327  &  87.53/96.35 \\
Seen-TI & 0.6700/0.7458  &  86.96/96.75       \\  \hline

\end{tabular}
\end{table}
\paragraph{Performances on Seen Target IDs and Source Faces}
In Tab.~\ref{tab:inout}, we present the performances of IDRetracor on the testing set of seen target IDs (Seen-TI), seen source faces (Seen-SF), and unseen data. See Fig.~\ref{fig:seennunseen} for the visualization and illustration of seen and unseen data. Notice that Seen-SF refers to the specific source faces that are applied to generate fake faces in the training set, and Unseen is the protocol we followed in all other experiments. The results show that there is minor variation in performance on seen and unseen data, which suggests that the abundant number of samples in the dataset has addressed the distinctions among Seen-TI, Seen-SF, and Unseen, thus achieving generalization. 

\subsection{Analysis of IDRetracor}

\begin{figure}[htbp]
	\centering
	\includegraphics[width=0.9\linewidth]{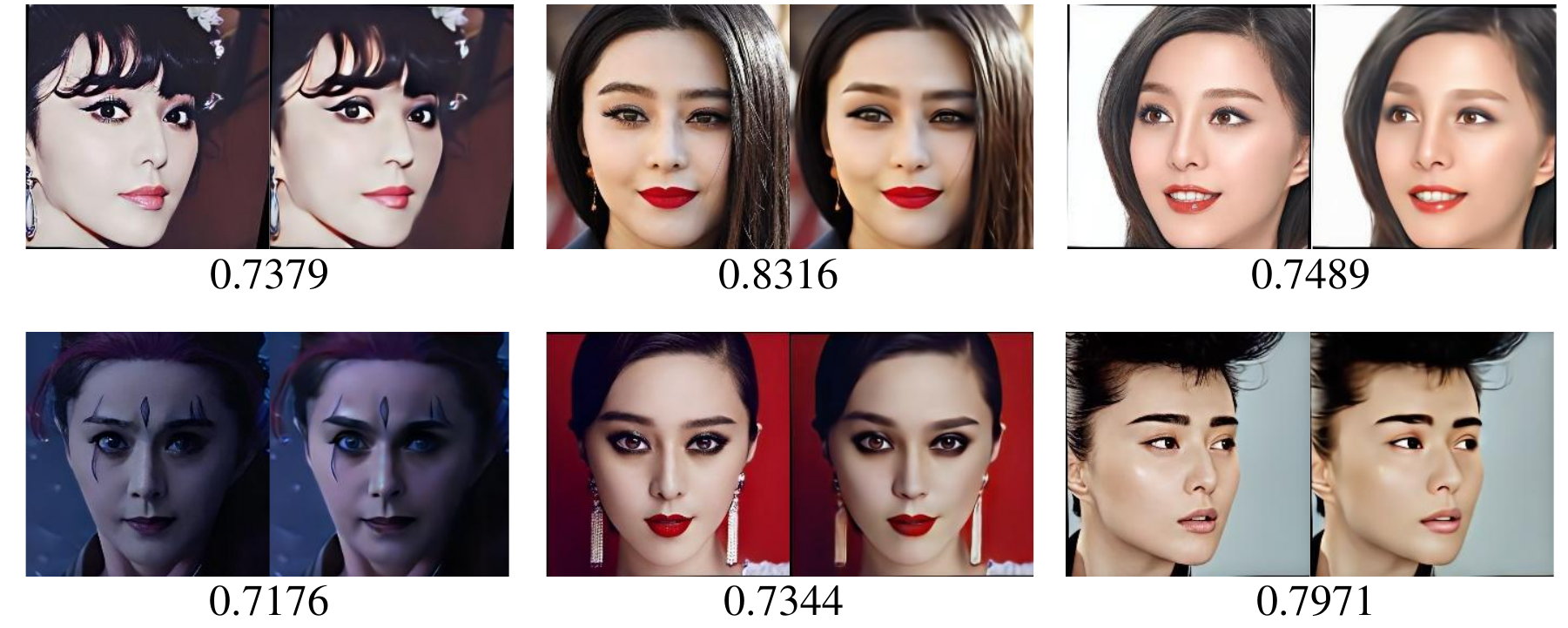}
	\caption{The retracing results when the inputs are real faces. The left and right faces of each pair are real and retraced faces, respectively. The values below each pair are their Arcface similarity scores. }
	\label{fig:s2s}
\end{figure}
\paragraph{Real Face Identity Preservation.}
In Fig. \ref{fig:s2s}, we substantiate the efficacy of incorporating source-source sample pairs into our model. For testing, we exclusively use source faces that are not seen during training. Qualitatively, the real-face identities are well-preserved in the output faces. We also show the Arcface similarity as the quantitative metric and the results are correspondingly promising. Therefore, the proposed IDRetracor is logically comprehensive for input reconstruction.

\begin{figure*}[h]
	\centering
	\includegraphics[width=0.7\linewidth]{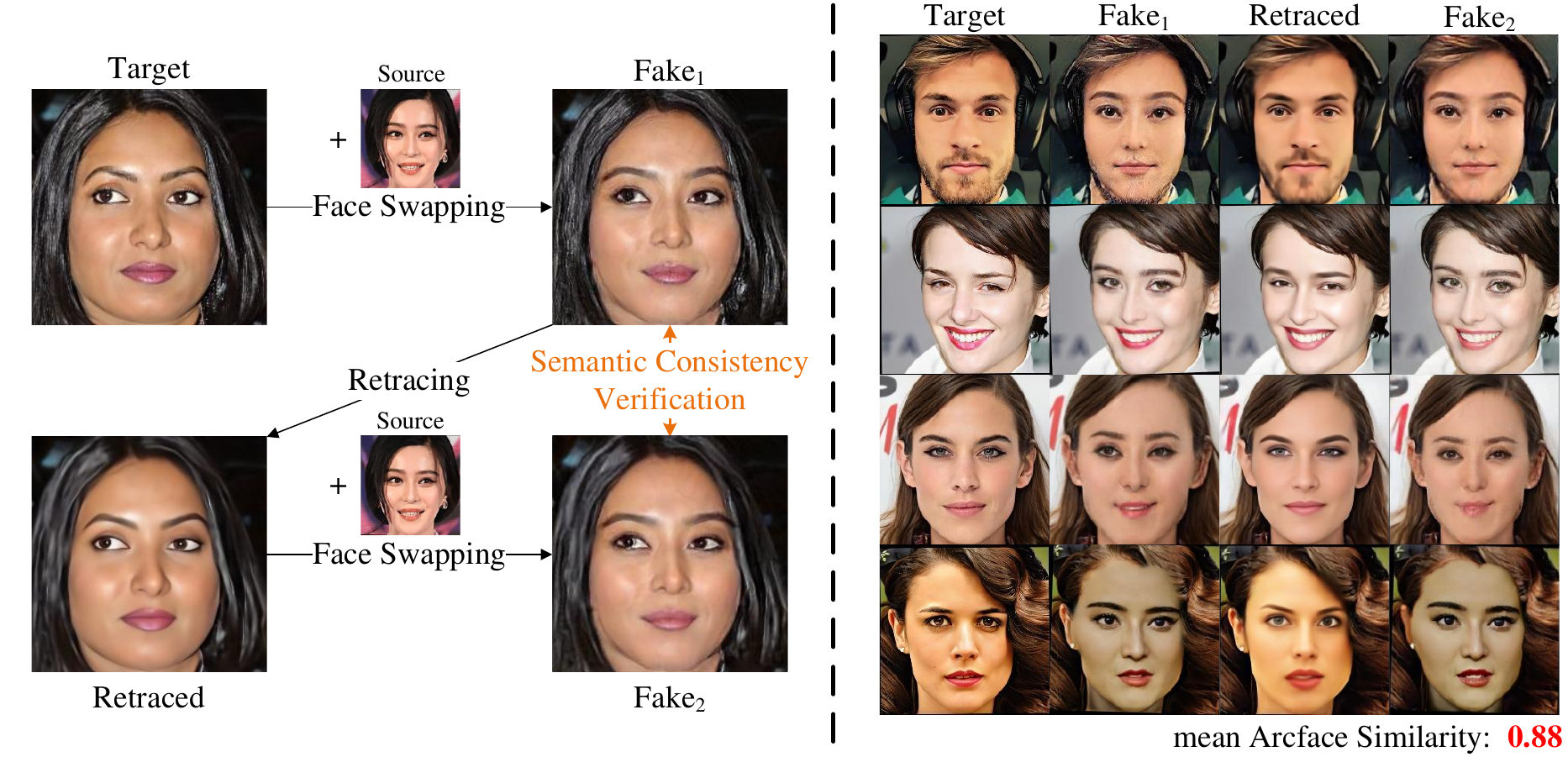}
	\caption{Left: Illustration for the process of semantic consistency verification. Right: Results of fake faces produced using the retraced target faces. mAS denotes the mean Arcface Similarity score between the original fake faces (Fake$_1$) and retraced fake faces (Fake$_2$) on the testing set. From top to bottom are SimSwap \cite{simswap}, InfoSwap \cite{infoswap}, HifiFace \cite{hififace}, and E4S \cite{e4s}, respectively.}
	\label{fig:semanticv}
\end{figure*}
\paragraph{Semantic Consistency Verification.}
To further assess the reliability of visual forensics, we design a semantic consistency verification experiment. Specifically, we randomly selected 1,000 fake faces from the test set and obtained their retraced faces. Subsequently, we perform face swapping between these retraced faces and their corresponding source faces used in the original fake face generation, yielding a set of new fake faces. If this set of new fake faces is consistent with the original fake faces, we can validate the semantic consistency of the retraced faces. As shown in Fig. \ref{fig:semanticv}, the experimental results verify the semantic consistency of the retraced faces.

\subsection{Results on Various Datasets}

\begin{figure}[htbp]
    \includegraphics[width=\linewidth]{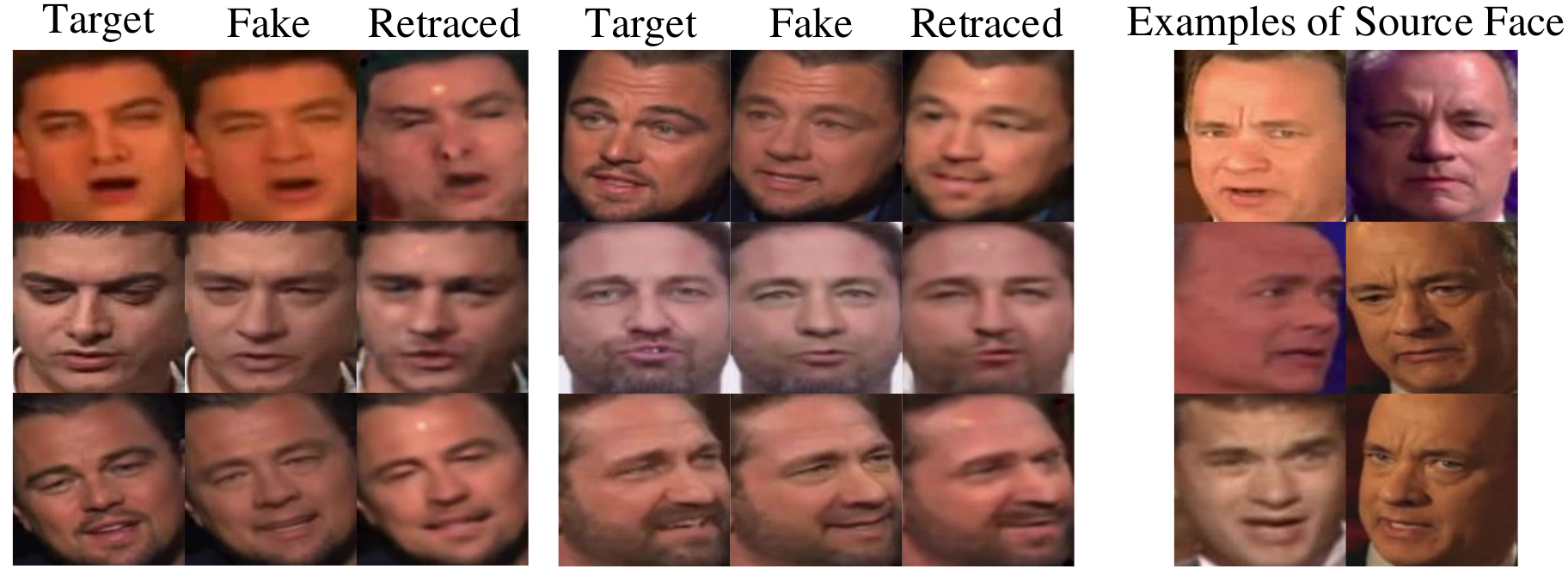}
    \caption{Examples of retracing fake faces in Celeb-DF-v2. Since the source face corresponding to the fake face is not specifically provided by the dataset, we present some examples of source identity on the right.}
    \label{fig:celeb}
\end{figure}

\begin{figure}[htbp]
\centering
    \includegraphics[width=0.7\linewidth]{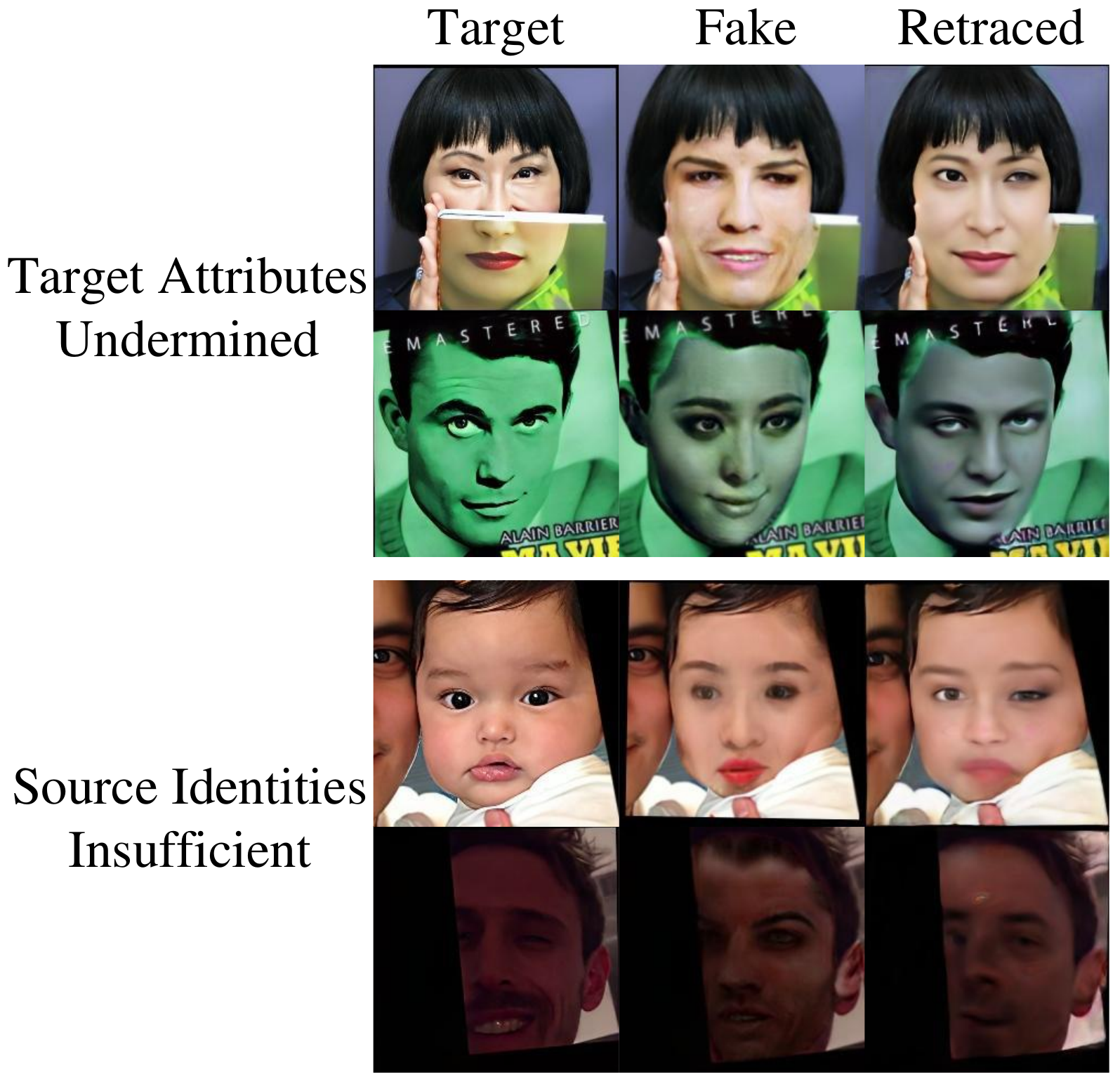}
    \caption{Failure cases of retracing.}
    \label{fig:fail}
\end{figure}
\paragraph{Training on Public Deepfake Datasets}
In Sec.~\ref{imp}, we mentioned that the two public datasets have too few target IDs, namely, a maximum of 26 target IDs for Celeb-DF-v2 \cite{Celeb} and a mere single one for FaceForensics++ \cite{Faceforensics}. Given that the target ID in FaceForensics++ is obviously insufficient for training a retracing network, we skip FaceForensics++ and conduct training experiments on Celeb-DF-v2 (see Fig. \ref{fig:celeb} ) following the same setting as on the crafted dataset. The poor experimental results show that we cannot train a retracing network on these public datasets, thus demonstrating the necessity of crafting our target-fake dataset. 
Moreover, it is also impractical to craft datasets applying the manipulation methods used by Celeb-DF-v2 or FaceForensics++. That is because they are traditional one-to-one methods that require training a new model for each target ID, while we need a massive number of target IDs in the dataset for retracing.

\paragraph{Cross-dataset Evaluation}

\begin{figure*}[htbp]
    \centering
    \includegraphics[width=0.8\linewidth]{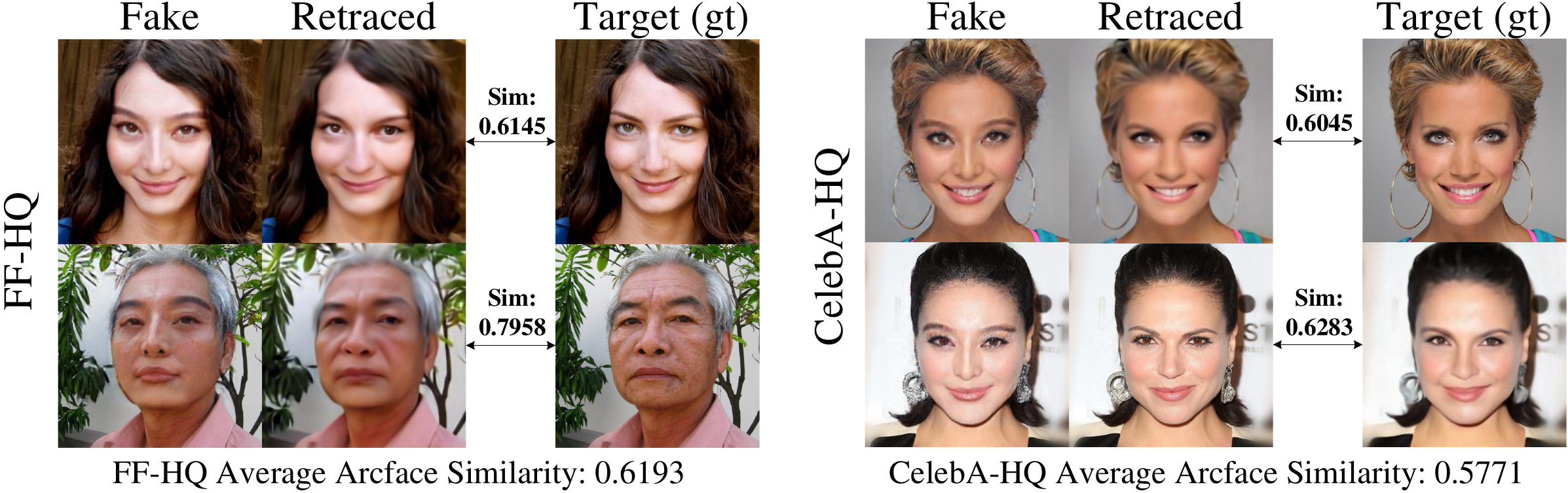}
    \caption{Cross-dataset Evaluation. }
    \label{fig:c-dataset}
\end{figure*}
Considering cross-dataset evaluation is important for real-world applications, we conduct the cross-dataset evaluation on FF-HQ \cite{FFHQ} and CelebA-HQ \cite{CelebAHQ} (trained on VGGFace2). The cross-dataset fake images are generated using target faces from FF-HQ and CelebA-HQ and exactly following the protocol that we apply for VGGFace2.
Fig. \ref{fig:c-dataset} shows that our method can maintain promising performance in the cross-dataset evaluation both qualitative and quantitative, which further demonstrates the application potential of IDRetracor. 

\begin{figure*}[h]
    \centering
	\begin{subfigure}[]{0.47\textwidth}
		\centering
		\includegraphics[width=\textwidth]{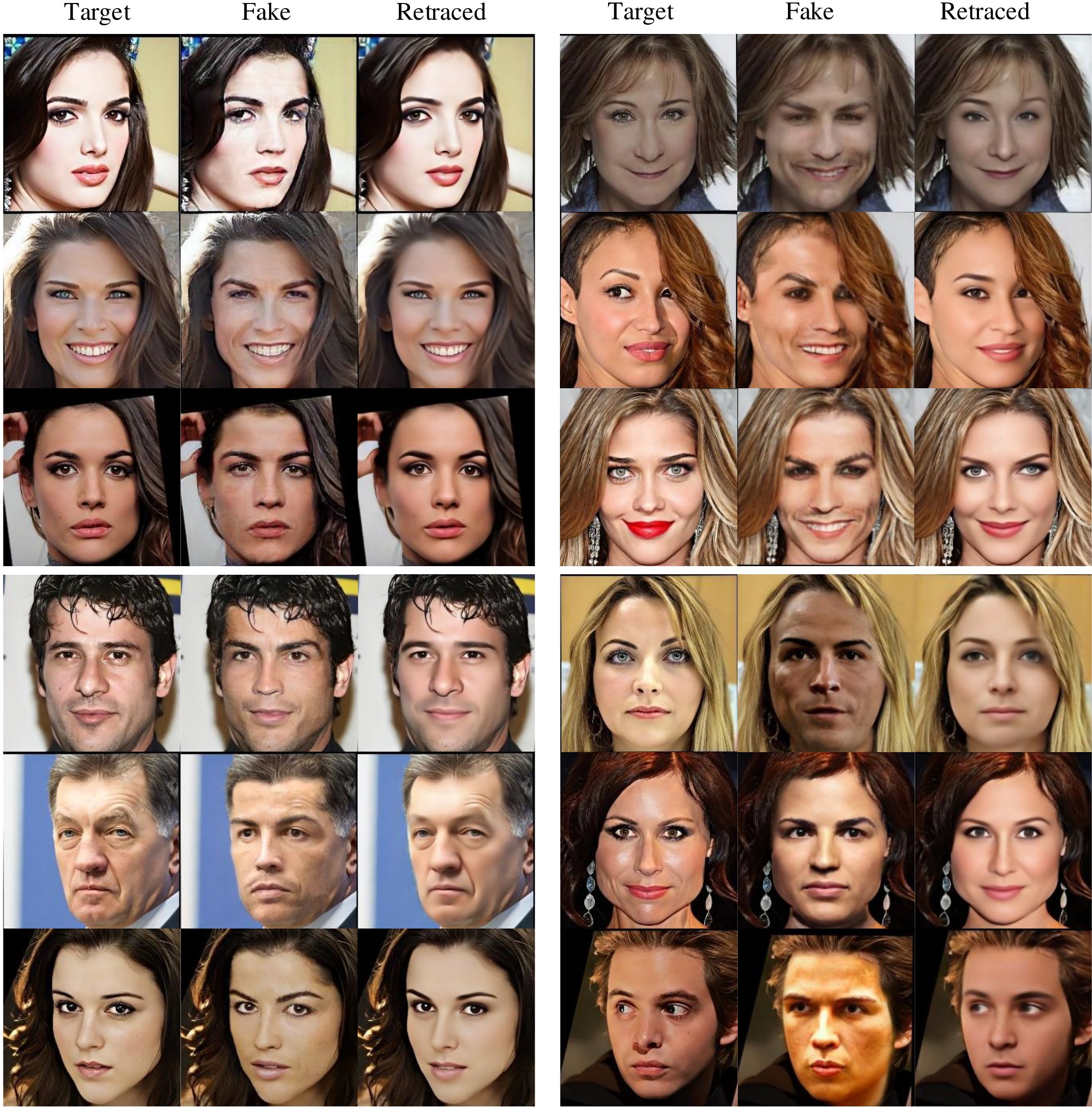}
		\caption{Ronaldo}\label{a_2}
	\end{subfigure}
	\begin{subfigure}[]{0.475\textwidth}
		\centering
		\includegraphics[width=\textwidth]{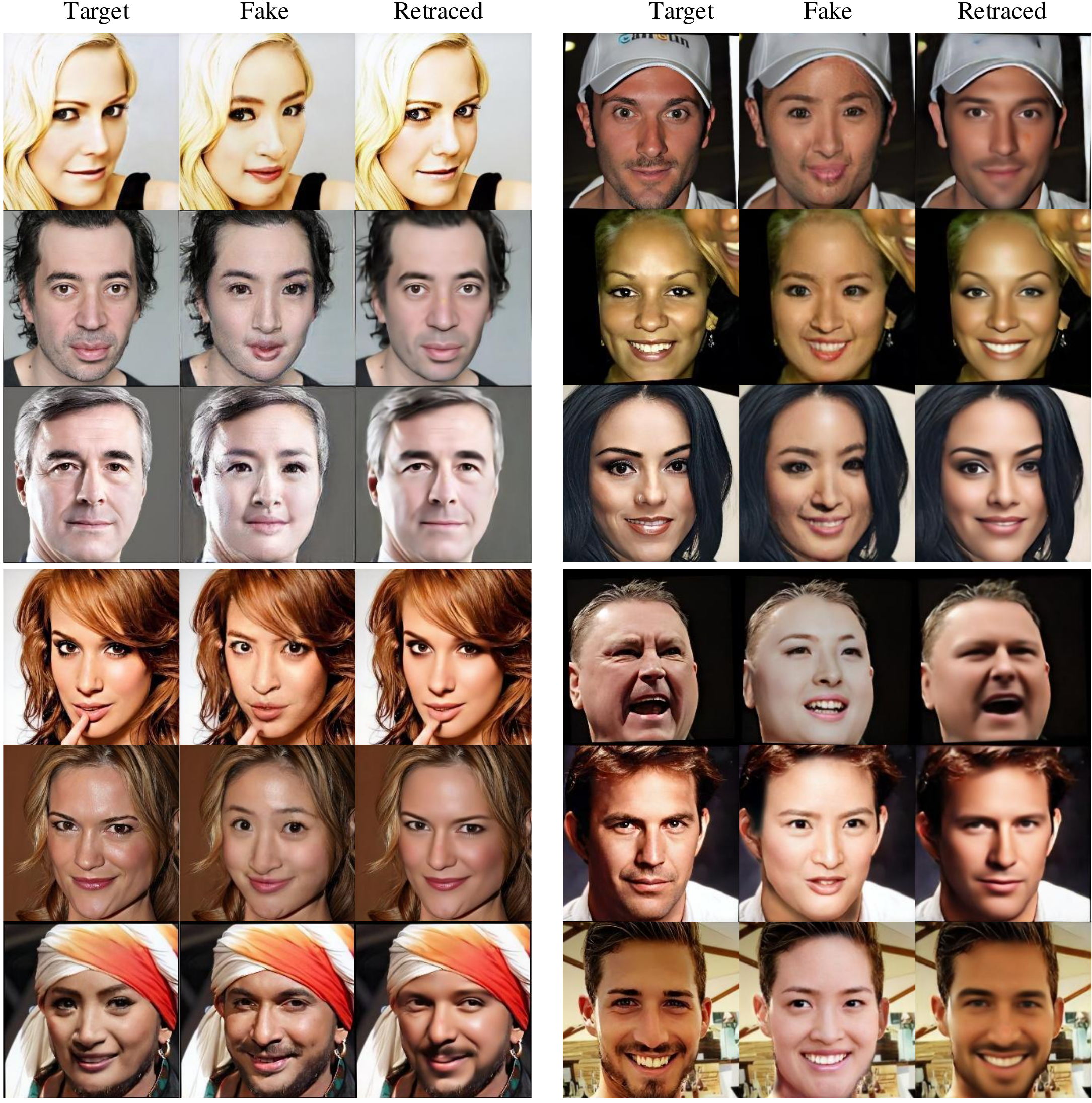}
		\caption{Lin}\label{b_2}
		
	\end{subfigure}
\caption{Qualitative results of retracing various source IDs. The fake faces are generated by Top-left: SimSwap, Top-right: InfoSwap, Bottom-left: HifiFace, and Bottom-right: E4S, respectively. Zoom in for better illustration.}\label{fig:RL}
\end{figure*}

\begin{table}[h]

	\centering
		\caption{The quantitative evaluations on two more source IDs.}\label{tab:main}
  \footnotesize
  \setlength{\tabcolsep}{10pt}
	\begin{tabular}{lcc}
		\hline
		Method & ID Sim. (arc/cos) & ID Ret. (t1/t5)   \\  \hline 
		Fake (Lin)       & 0.1319/0.1770  &   6.15/10.90 \\ 
		Ours (Lin)     & 0.6691/0.7110  &     87.13/96.05  \\ \hline
		Fake (Ronaldo)       & 0.1019/0.1585  &  5.73/8.95 \\ 
		Ours (Ronaldo) & 0.6474/0.6998  &   86.43/93.15 \\ \hline
	\end{tabular}
\end{table}

\paragraph{Retracing Various Source IDs}
Here, we demonstrate that the framework of IDRetracor can be applied to other source IDs effectively. We craft datasets and train the corresponding IDRetracors for two other celebrities (\textit{i.e.}, Ariel Lin and Cristiano Ronaldo from VGGFace2). As shown in Fig.~\ref{fig:RL}, the IDRetracor can also exhibit strong retracing performance for these two source IDs, where the identity and detailed facial characters are promisingly reconstructed. 
In Tab.~\ref{tab:main}, we also provide the quantitative metrics on these identities.
Due to the varying performances of face swapping methods when dealing with different source faces, there are minor differences in the retracing results on quantitative metrics. Nonetheless, IDRetracor can still maintain effective retracing performances, indicating that our framework can be applied to any source ID.

\section{Failure Cases, Limitation, and Future Works}
As shown in Fig. \ref{fig:fail}, we provide some faces that the IDRetracor fails to retrace. Since the learned feature of IDRetracor is majorly dependent on the residual artifacts and the specific source identity, two types of fake faces are challenging to be retraced: 1) The target attributes are undermined after face swapping (see the first and the second rows). 2) The fake faces share less similarity with the source faces, that is, the violation of Premise 2 (see the third and the fourth rows).

The dependence on a specific source ID influences the practical utility of IDRetracor. This leads to the requirement to rebuild the target-fake dataset and retrain the model for each unique source subject, thus incurring extra time consumption.

In future works, it is vital to first conduct iterative experiments to find a backbone network most suitable for retracing tasks. Then, finding an efficient way to incorporate source face adaptively based on Premise 2 is crucial for the real-world application of IDRetracor.

\section{Conclusion}
In this paper, we analyze the urgent toward visual forensics and introduce the face retracing task to address it. Firstly, we discuss the feasibility and premise of the retracing task. Then, we propose a framework named IDRetracor for the retracing task, which can locate the solution space of the original target face and reconstruct the retraced face with the correct ID. Finally, extensive experiments demonstrate that the IDRetracor can retrace fake faces of arbitrary target IDs generated by multiple face swapping methods.\\

{
\small
\bibliographystyle{plain}
\bibliography{refer}
}

\end{document}